\documentclass[runningheads]{llncs}
\usepackage{graphicx}
\usepackage{subcaption}
\newcommand\tab[1][1cm]{\hspace*{#1}}

\begin{document}
\title{Automatic Airway Segmentation in chest CT using Convolutional Neural Networks}
\author{Garcia-Uceda Juarez, A.\inst{1,2}, Tiddens, H.A.W.M.\inst{2,3}, de Bruijne, M.\inst{1,4}}
%
%\titlerunning{Automatic Airway Segmentation in chest CTs using CNNs}
%
%\authorrunning{Garcia-Uceda Juarez et al.}
%
\institute{Biomedical Imaging Group Rotterdam, Departments of Medical Informatics and Radiology, Erasmus MC, 3015 CE, Rotterdam, The Netherlands \\ \email{a.garciauceda@erasmusmc.nl} \and Department of Pediatric Pulmonology, Erasmus MC-Sophia Children Hospital, 3015 CE, Rotterdam, The Netherlands \and Department of Radiology and Nuclear Medicine, Erasmus MC-Sophia Children Hospital, 3015 CE, Rotterdam, The Netherlands \and Department of Computer Science, University of Copenhagen, DK-2100, Copenhagen, Denmark}
\maketitle              % typeset the header of the contribution
\begin{abstract}
Segmentation of the airway tree from chest computed tomography (CT) images is critical for quantitative assessment of airway diseases including bronchiectasis and chronic obstructive pulmonary disease (COPD). However, obtaining an accurate segmentation of airways from CT scans is difficult due to the high complexity of airway structures. Recently, deep convolutional neural networks (CNNs) have become the state-of-the-art for many segmentation tasks, and in particular the so-called Unet architecture for biomedical images. However, its application to the segmentation of airways still remains a challenging task. This work presents a simple but robust approach based on a 3D Unet to perform segmentation of airways from chest CTs. The method is trained on a dataset composed of 12 CTs, and tested on another 6 CTs. We evaluate the influence of different loss functions and data augmentation techniques, and reach an average dice coefficient of 0.8 between the ground-truth and our automated segmentations.
\end{abstract}
\section{Introduction}
Segmentation of airways in chest computed tomography (CT) images is critical to obtain reliable biomarkers to assess the presence and extent of airway diseases. Biomarkers such as airway lumen diameter, airway wall thickness, airway tapering and airway-artery diameter ratio help in detection of early signs of disease and quantification of its severity. However, the segmentation of airways is a difficult task due to the complex tree-like structure of airways, with a large number of branches of very different sizes and orientations.

There are several methods proposed for automatic extraction / segmentation of the airway tree in chest CTs. Conventional methods such as region growing can successfully capture the main bronchi, but systematically fail at extracting the peripheral bronchi of smaller size. Other more sophisticated techniques are:~\cite{ref1}, based on a region growing approach using an airway probability map computed by a voxel classifier, together with an airway / vessel orientation similarity term;~\cite{ref2}, based on a tube-likeness shape detector computed from the Gradient Vector Flow field properties; or~\cite{ref3,ref4}, based on optimal surface graph-cut methods which find a globally optimal solution accounting for both airway lumen and outer wall surfaces with a wide range of tailored geometric constraints. These and other airway segmentation methods were evaluated by Lo et al. in the EXACT09 challenge~\cite{ref5}.

Since recently, the state-of-the-art methods for many image segmentation tasks are based on convolutional neural networks (CNNs)~\cite{ref6}. In particular, the so-called Unet network proposed by Ronneberger et al.~\cite{ref7} has become popular in segmentation tasks of biomedical images. With regards to airway segmentation, a number of CNNs-based methods~\cite{ref8,ref9,ref10,ref11} have been proposed which have outperformed the “classical” methods compared in~\cite{ref5}. In particular,~\cite{ref8,ref9,ref10} use the 3D Unet network to analyse volumetric images. Another novel method for airway extraction is proposed by Selvan et al.~\cite{ref12}, by formulating it as a mean field approximation based graph refinement task that resembles feed forward neural networks.

In this paper, we propose a robust fully automatic end-to-end method based on 3D Unets to perform airway segmentation in chest CT images. Other Unet-based approaches are more complex, such as~\cite{ref8} which relies on a tracking algorithm of the connected structure of the airway tree, and uses a local volume of interest (VOI) around single tracked branches. In contrast, our method is simpler and end-to-end optimised, whose only input are large images patches containing various branches at once. This makes it more robust.
%
%
%\vspace{-0.2cm}
\section{Methodology}
The processing pipeline for airway segmentation proposed in this work is described in the next subsections, including pre- and postprocessing techniques.
%
%\vspace{-0.2cm}
\subsection{Network architecture}
The 3D Unet is adopted for volumetric image analysis by replacing the operators of the original 2D U-Net proposed in~\cite{ref7}: i) convolution; ii) pooling; iii) upsampling; with their analogous 3D operator. The Unet method consists of an encoder / downsampling path followed by a decoder / upsampling path, each with 5 levels of resolution. Each level in the downsampling path is composed of two 3x3x3 convolutional layers followed by a 2x2x2 pooling layer. At each level the number of feature channels is doubled, with a number of channels in the first level of 16. In the upsampling path, only one 3x3x3 convolutional layer is used before an 2x2x2 upsample layer. This is in order to reduce memory requirements of the network, since convolutions in the upsampling path can be less relevant, as mentioned in~\cite{ref13}. Each convolution operator is followed by a rectified linear unit (ReLU). At each level the number of feature channels is halved. The final layer consists of a 1x1x1 convolutional layer followed by a sigmoid activation function. This way the network output is a voxelwise probability map of the sought airways, of the same size as input images. The total number of convolutional layers in the network is 15.

In the convolution operations, an adequate zero-padding is used to obtain an output of the same size as the input. This is for sake of simplicity in designing  the same network for arbitrary sizes of input image. Moreover, due to size constraints of the input images, we disable the convolution / pooling operators applied in axial direction in the deepest level of the network, i.e. we use 1x3x3 convolutions and 1x2x2 pooling.

As regularisation, we use exclusively on-the-fly data augmentation during training, explained in section 2.3. No dropout has been used. The reason for this is twofold: i) our experiments showed that data augmentation was more efficient for regularisation, and ii) the use of dropout in Keras incurs in a large increase in memory footprint, since the feature maps input to the dropout layers are duplicated. Indeed, the main challenge of our experiments is the maximum size of the network that fits in GPU memory, more important than computational speed. And while the inclusion dropout layers requires a significant reduction of size, data augmentation is generated on the fly with no memory overhead and a negligible increase in computational time. 
%
%\vspace{-0.2cm}
\subsection{Choice of loss function}
Two different loss functions for training the network have been tested in our experiments, namely: i) weighted binary cross-entropy (wBCE) (Eq. 1), and ii) Dice coefficient (dice) loss (Eq. 2). 

\begin{equation}
\mathcal{L}_1 = w_B \sum_{x\in N_B \cap N_L}{\log(1 - p(x))} + w_A \sum_{x\in N_A \cap N_L}{\log(p(x))} 
\end{equation}
\begin{equation}
\mathcal{L}_2 = \frac{2 \sum_{x\in N_L}{p(x) g(x)}}{\sum_{x\in N_L}{p(x)} + \sum_{x\in N_L}{g(x)} + \epsilon} 
\end{equation}
Where $p(x), g(x)$ are the voxelwise airway probability maps and airway ground-truth, respectively. The subindexes $B, A$ refer to the ground-truth classes background / airways, respectively, and $L$ corresponds to the region of interest (ROI): the lungs. $N_k$ is the group of voxels for each class. $\epsilon$ is a tolerance needed when there are no ground-truth voxels found in the (sub)image.

We force the training of the network to the ROI: the lungs fields, so that only voxels within this region contribute to the loss. This is achieved by masking the probability maps and ground-truth with a lung segmentation. This is indicated by the intersection $N_B \cap N_L$ in Eqs. 1-2. This approach forces the voxelwise classification to focus only on discriminating between airways and lung parenchyma (background), ignoring other non-relevant structures including ribs, adipose tissue, and skin. The lung segmentation needed in this approach is easily computed by a region-growing method~\cite{ref1}.

In the wBCE function (Eq.1), the weights $w_B, w_A$ are used to compensate for a large interclass imbalance of lung parenchyma (background) / airways voxels. The weights are computed on the fly, for a given input image patch, as the ratio of the opposite class voxels with respect to the  the total voxels, or analogously: $w_B = 1, w_A = N_B / N_A$.
%
%\vspace{-0.2cm}
\subsection{Preprocessing}
The main challenge of our experiments is that chest CT images have a size much larger than the maximum input to the network that fits in GPU memory. We apply several preprocessing steps to adapt the input images and reduce the memory footprint.
\subsubsection{Cropping images}
The CTs are cropped to the region of interest: the lungs. The bounding-box is of fixed size in x,y dimensions: 352x240 pixels, and centered in each lung. The axial dimension is different for each CT. The box is enlarged by 30 voxels in each direction to account for border effects.
\subsubsection{Sliding-window}
A sliding-window approach is used to extract smaller image patches from the input CTs that fit to the size requirements of the network, similar to the method in~\cite{ref6}. The sliding is undertaken only in axial direction with a stride of 104 pixels, while the x,y dimensions are fixed and equal to the network input size: 352x240 pixels. This is schematically shown in Fig.~\ref{fig1}. The sliding stride is set to have 75\% overlap in between the extracted patches, in order to prevent border effects. The sliding-window is used for all CTs of training and validation sets. The resulting image patches are generated on the fly during training, to prevent any memory overhead, and are randomly shuffled at the beginning of each epoch.
\subsubsection{Data augmentation}
\begin{figure}
\centering
\includegraphics[width=\textwidth]{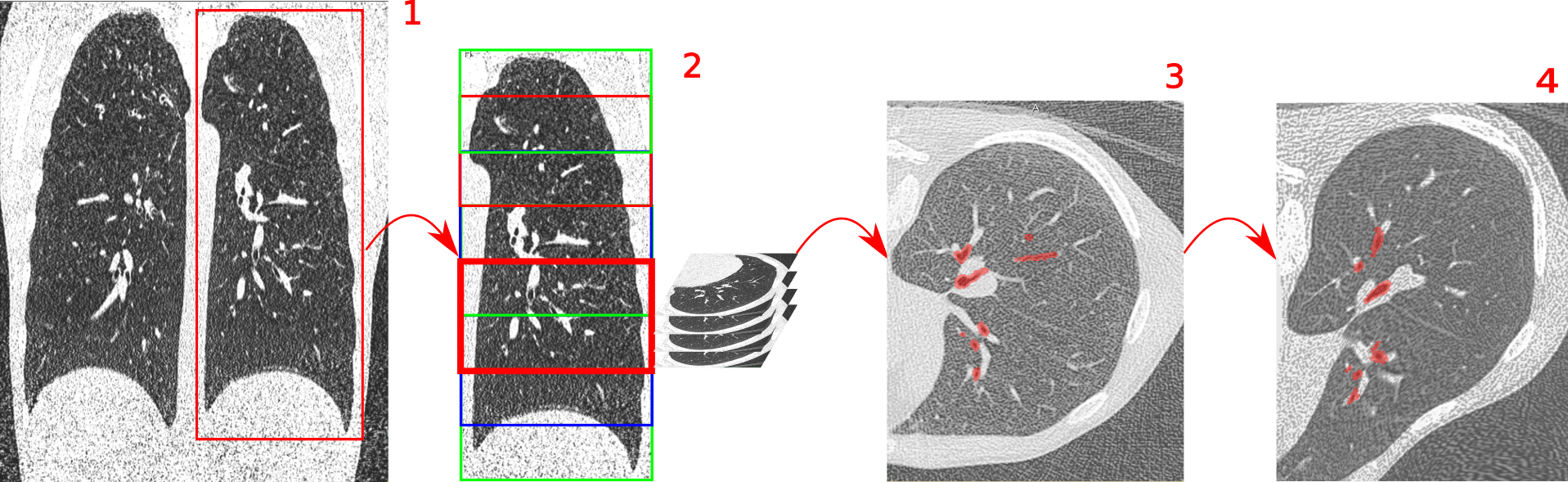}
\caption{Preprocessing pipeline of input images to the network. 1: crop CTs. 2: Extract 3D images patches by sliding-window. 3,4: Apply data augmentation (elastic deformations) to the input patch, which is fed to the network.} \label{fig1}
\end{figure}

Data augmentation is applied systematically to the input images during training, by means of i) rigid transformations, and/or ii) elastic deformations. The former operations consist of random flipping and rotations with a maximum value of 10 degrees, in the three dimensions. The elastic deformations consist of smooth voxel displacements computed using bicubic interpolation from random displacement vectors on a coarse grid of 3x3, which are sampled from a Gaussian distribution with 25 voxels standard deviation. This methodology is similar to the one used in~\cite{ref7}. This is schematically shown in Fig.~\ref{fig1}. The data augmentation is applied over the 3D images patches extracted by the sliding-window method, on both the training and validations sets. The resulting images are generated on the fly during training, and these are the input to the network.
%
%\vspace{-0.2cm}
\subsection{Post-processing}
Zero-padding in convolution operations in the network is used for simplicity, but the results suffer from border effects where the probability maps are less reliable towards the image boundaries. Instead, a network composed of valid convolutions ensures a fully reliable output, but its size is strongly reduced to roughly 25-50\% of input size. This is an issue in our experiments where the axial dimension of input patches is rather small.

\begin{figure}
\centering
\includegraphics[width=\textwidth]{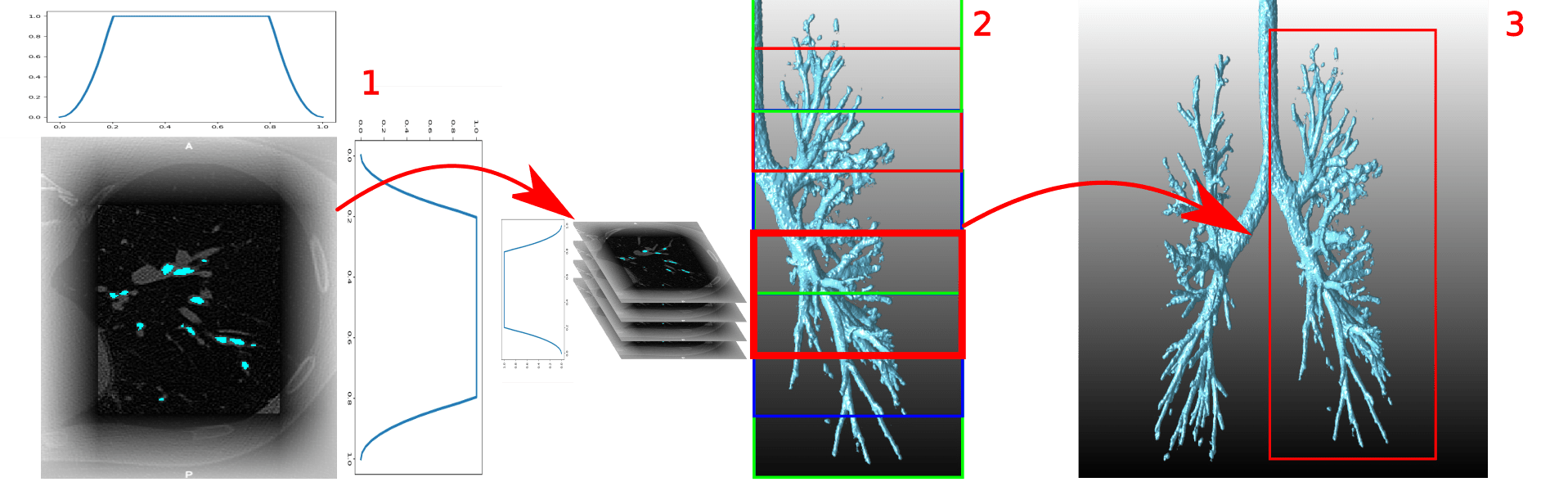}
\caption{Postprocessing of output probability maps. 1: diminish output near image boundaries to prevent border effects. 2,3: Reconstruct full size probability maps, and apply thresholding to obtain airway tree.} \label{fig2}
\end{figure}

The technique implemented to prevent border effects is schematically shown in Fig.~\ref{fig2} and works as follows: the output probability maps are multiplied with a function that is: i) "1" within a window corresponding to the output of a similar network with valid convolutions, ii) elsewhere, a quadratic polynomial decreasing towards the output borders. The function in 1D corresponds to Eq. 3, with $x_l, x_r$ the limits of the output in i), and image dimensions $x\in[0, x_m]$.

\begin{equation}
f(x) =
\left\{
	\begin{array}{ll}
		(x/x_l)^2  & \mbox{if } x < x_l \\
		1 & \mbox{if } x_l \leq x \leq x_r \\
		((x_m-x)/(x_m-x_r))^2 & \mbox{if } x > x_r \\
	\end{array}
\right.
\end{equation}
The full size output is reconstructed by placing together output patches following the structure generated for input images, and normalizing for patches overlap. Finally, the output outside the lung ROI is discarded by masking the probability maps with the binary mask of lungs. The final airway tree is obtained by thresholding the resulting airway probability maps.
%
%
%\vspace{-0.2cm}
\section{Dataset and Experiments set-up}
The dataset used to conduct the experiments consists of 24 inspiratory chest CT scans from pediatric subjects, 12 controls and 12 with respiratory disease: 11 with Cystic fibrosis (CF) and 1 with Common Variable Inmune Deficiency (CVID). Both groups were gender and age matched: range 6 to 17 years old, 5 females, per group. Scanning was undertaken using spirometry-guidance in a Siemens SOMATOM Definition Flash scanner. Similar kernel reconstructions were used for all scans: I70f/3, B75f, and B70f. This dataset has been prepared in the works of~\cite{ref14,ref15}.

Each CT scan consists of a number of cross-sectional slices, from 265 to 971 slices, with varying thickness in the range 0.75-1.0 mm, and slice spacing from 0.3 to 1.0 mm. Each slice is of fixed size equal to 512x512 pixels, with a pixel size in the range 0.44-0.71 mm.

The 12 CTs for control and disease patients are randomly split in three categories: training, validation and testing, with proportion 50 / 25 / 25 \%, respectively. The final data groups are then: 12 CTs for training, 6 CTs for validation and 6 CTs for testing. 

The ground-truth used to train the network are reference segmentations of the airways outer wall obtained from an accurate airway segmentation method~\cite{ref4} applied on manual annotations of centrelines. These are manually extracted for the entire airway tree using specialised software in the work of~\cite{ref14}. A coarse segmentation is generated from the centrelines, and then the surface graph-cut method~\cite{ref4} is applied to refine this and obtain an accurate segmentation of both airway lumen and outer wall.

The experiments are conducted in a GPU NVIDIA GeForce GTX 1080 Ti with 11 GB memory. The network architecture has been implemented in Keras framework~\footnote{https://keras.io/}.
%
%\vspace{-0.2cm}
\subsection{Network Optimisation}
The network is designed to accommodate the largest input images possible that fit the GPU memory during training. This corresponds to input images of size 104x352x240 with a batch containing only one image. It has been observed that this is advantageous over increasing the batch size using smaller input images.

For training the network, the Adam optimizer is used with a learning rate of 1.0e-05. The training time is stopped either when i) the validation loss increases over 15 epochs, or ii) a maximum of 300 epochs is reached. In either case, the results consists of the model with the lowest validation error during training, evaluated on the test set. Training time is approximately 1 day, while testing time is only a few seconds per CT scan.
%
%\vspace{-0.2cm}
\subsection{Experiments set-up}
The various experiments conducted correspond to different set-ups of the network, namely: i) use of a) dice or b) wBCE loss function, and ii) use of a) rigid transformations or b) elastic deformations as data augmentation. All the experiments are displayed in Table~\ref{tab1}.
\begin{table}
\caption{List of set-ups of experiments conducted} \label{tab1}
\centering
\begin{tabular}{|l|l|l|l|}
\hline
\tab & Loss function & Data augmentation & Name \tab \tab\\
\hline
1 & wBCE & None & wBCE-None\\
2 & wBCE & Rigid & wBCE-Rigid\\
3 & dice & None & dice-None\\
4 & dice & Rigid & dice-Rigid\\
5 & dice & Elastic & dice-Elastic\\
\hline
\end{tabular}
\end{table}
\vspace{-0.2cm}
\begin{figure}[h]
\centering
\includegraphics[width=0.49\linewidth]{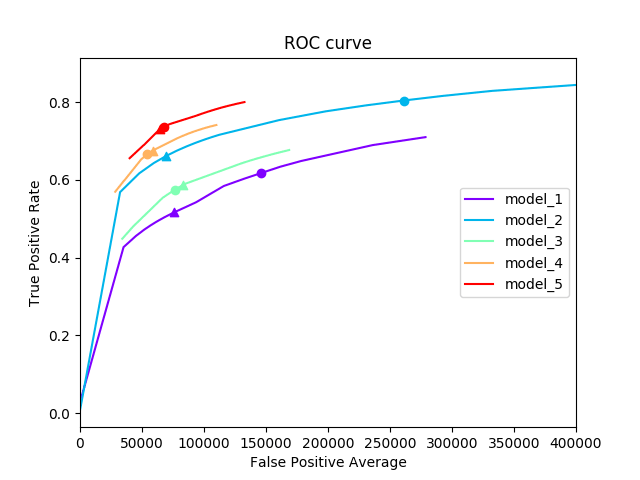}
\includegraphics[width=0.49\linewidth]{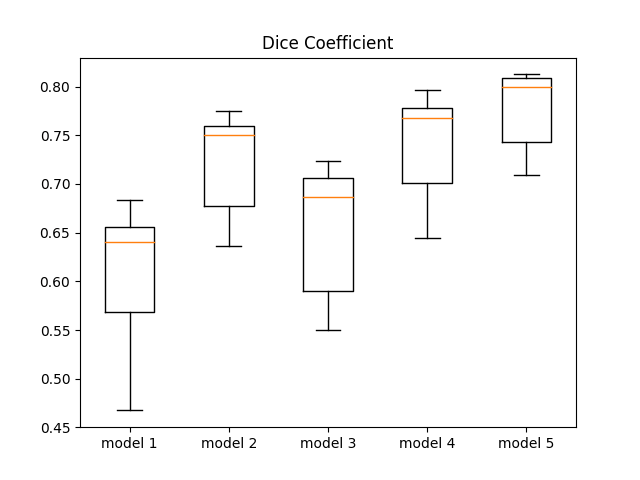}
\caption{Left: FROC curve computed on the test set, by varying the thresholding in the probability maps, for all models in Table~\ref{tab1}. The circle / triangle correponds to a threshold of 0.5 / optimal value, respectively. The optimal thresholds are: with dice loss: 0.5; with wBCE loss: around 0.9. Right: average Dice coefficient on the test set. The results correspond to the optimal threshold for each model.} \label{fig3}
\end{figure}
%
%
%\vspace{-0.2cm}
\section{Results and Discussion}

\begin{figure}[h]
\centering
\includegraphics[width=0.49\linewidth]{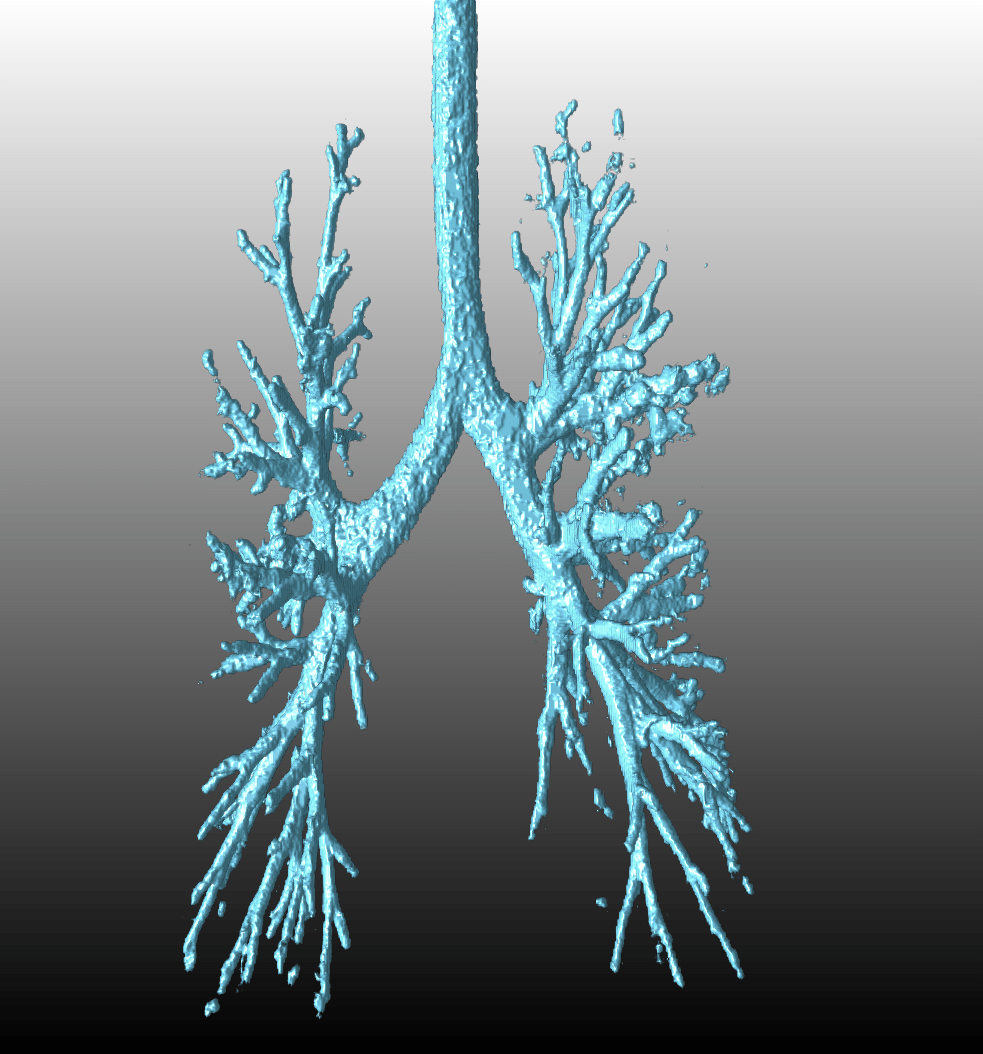}
\includegraphics[width=0.49\linewidth]{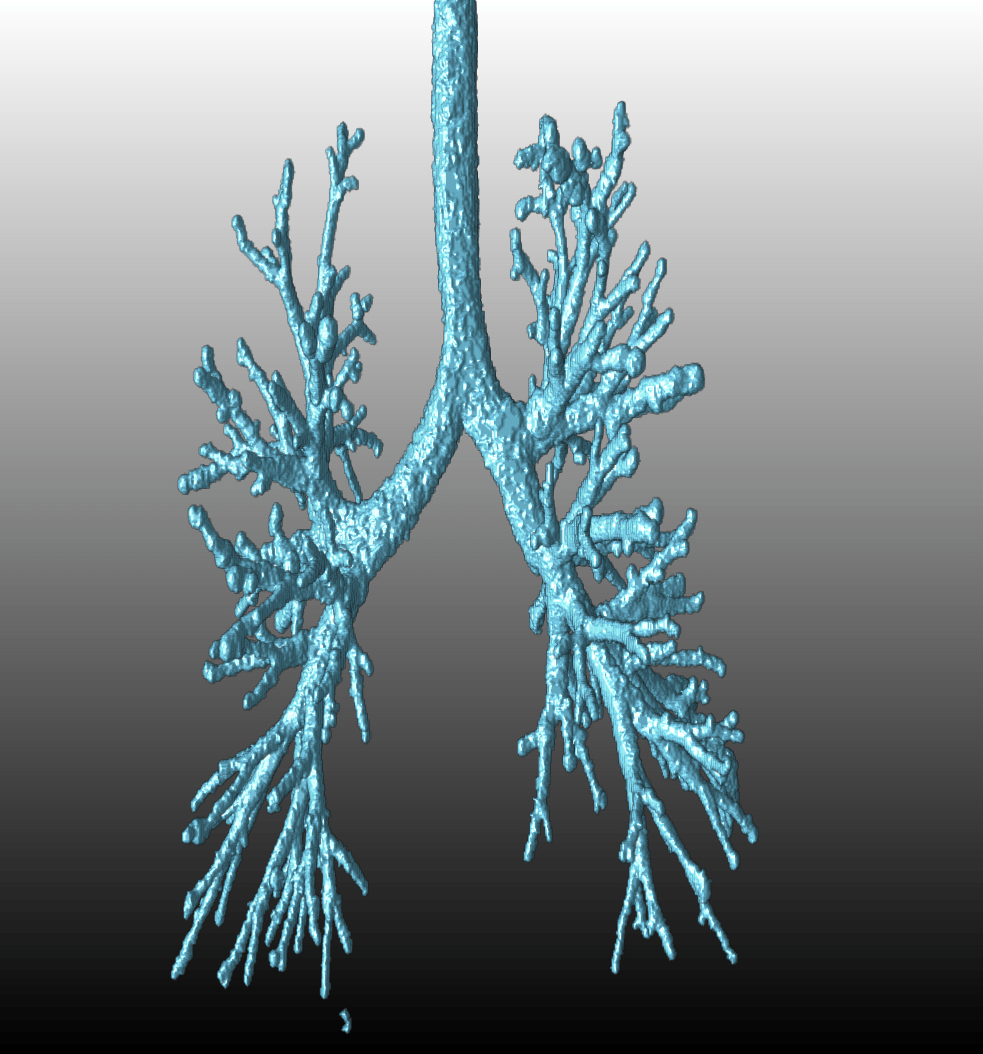}
\caption{Left: visualization of the segmented airway tree, obtained for one of test CTs with the model "dice-Elastic" and optimal threshold. Right: ground-truth} \label{fig4}
\end{figure}
\begin{figure}
\centering
\begin{subfigure}{0.48\textwidth}
\includegraphics[width=\linewidth]{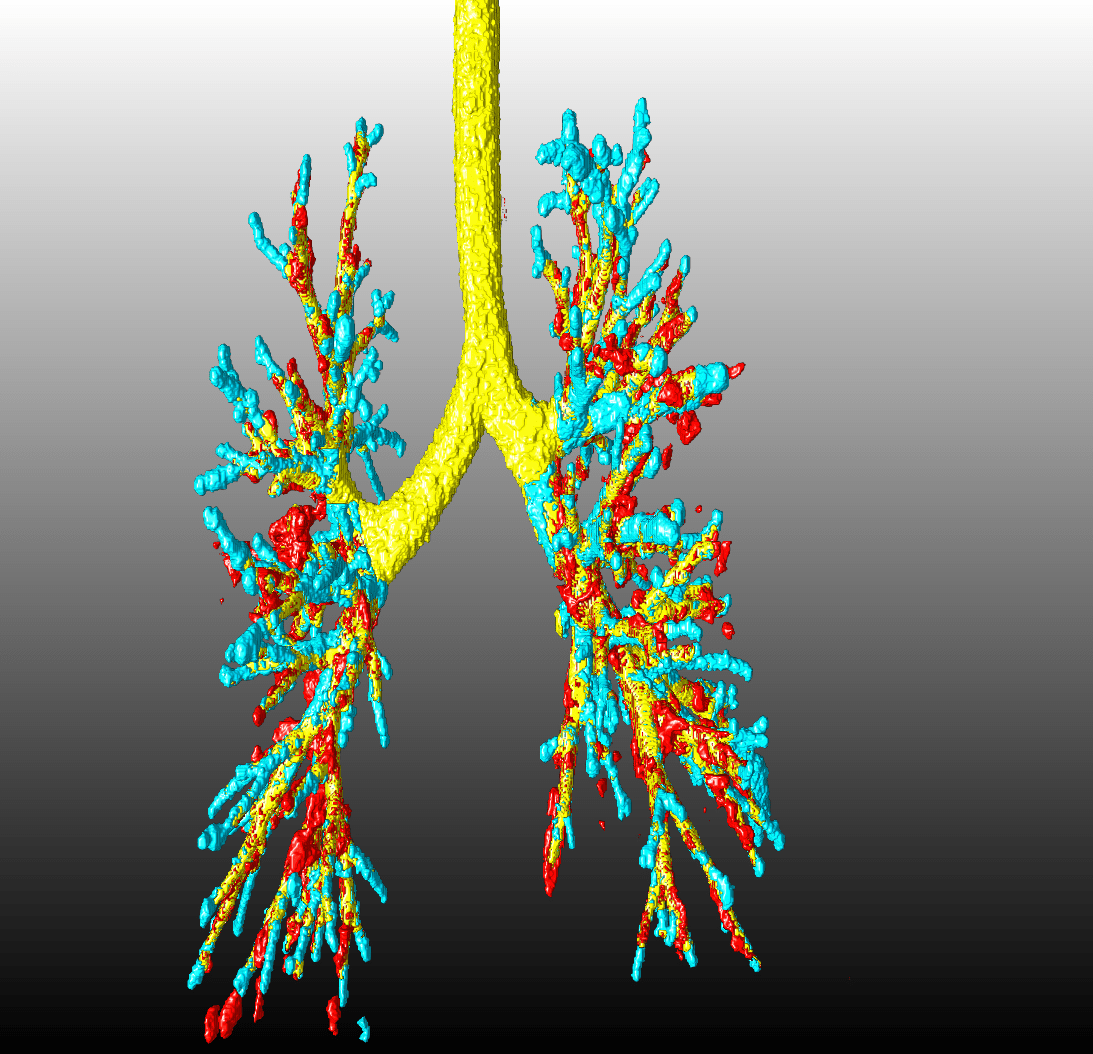}
\caption{wBCE-None}
\end{subfigure}
\centering
\begin{subfigure}{0.48\textwidth}
\includegraphics[width=\linewidth]{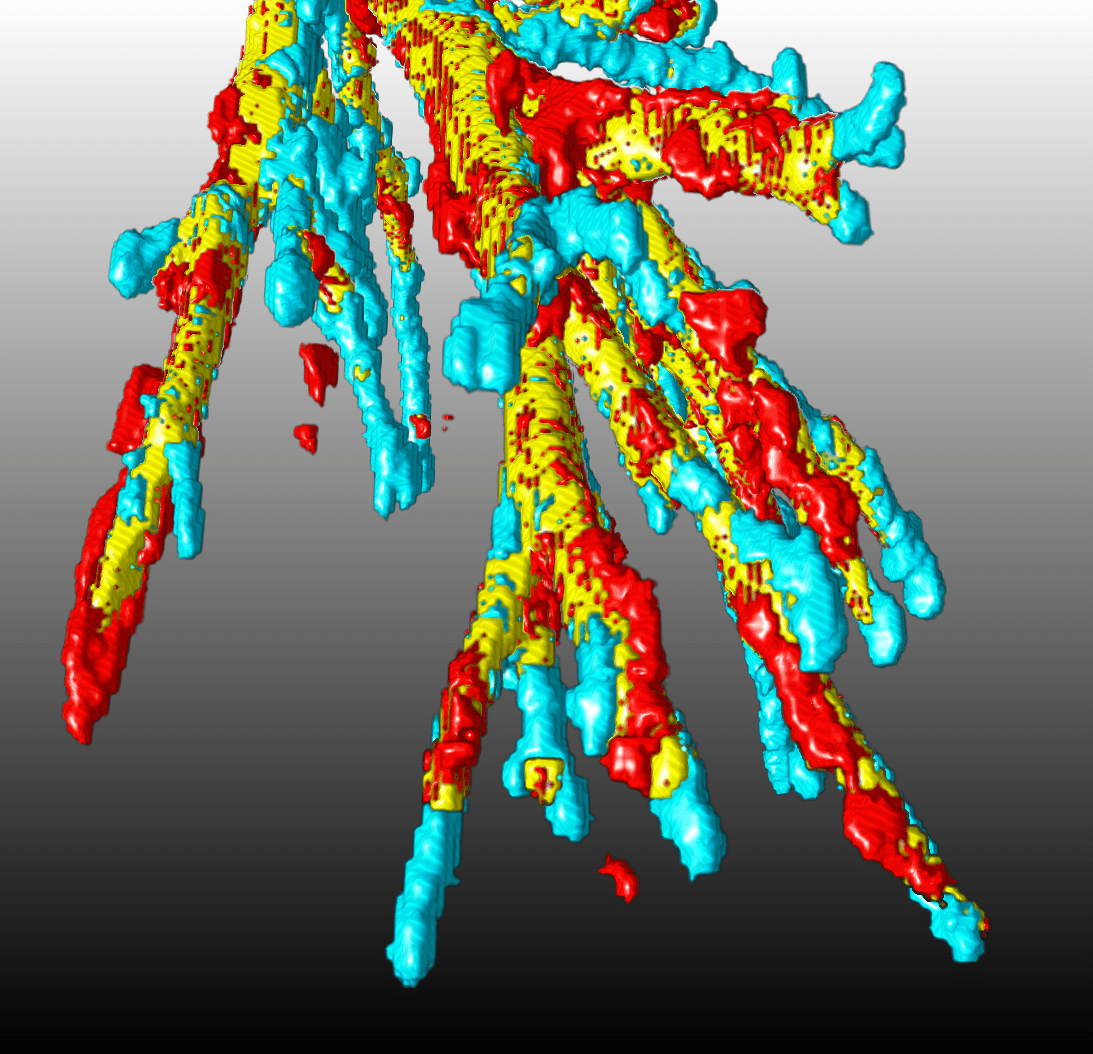}
\caption{wBCE-None, detailed view}
\end{subfigure}
\centering
\begin{subfigure}{0.48\textwidth}
\includegraphics[width=\linewidth]{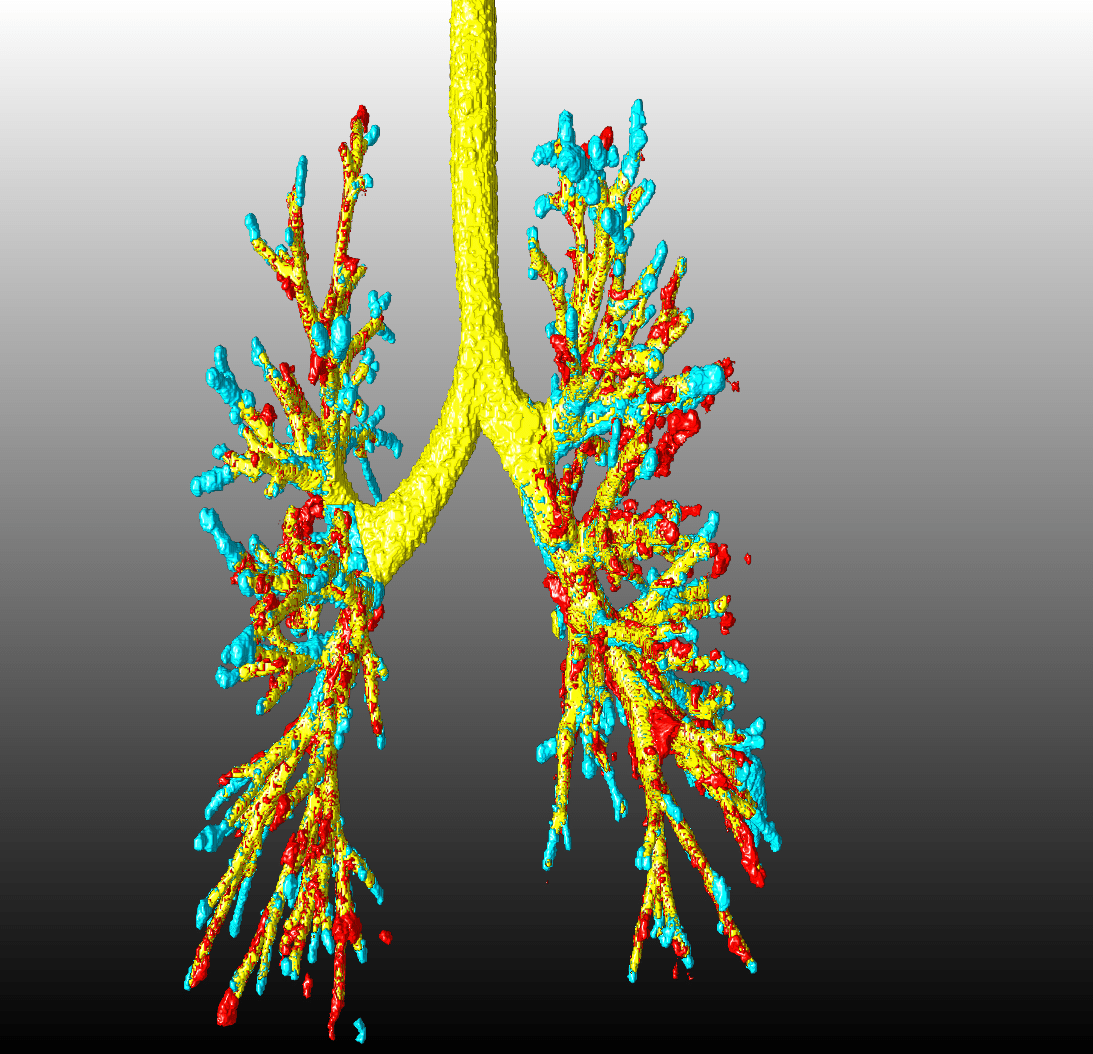}
\caption{wBCE-Rigid}
\end{subfigure}
\centering
\begin{subfigure}{0.48\textwidth}
\includegraphics[width=\linewidth]{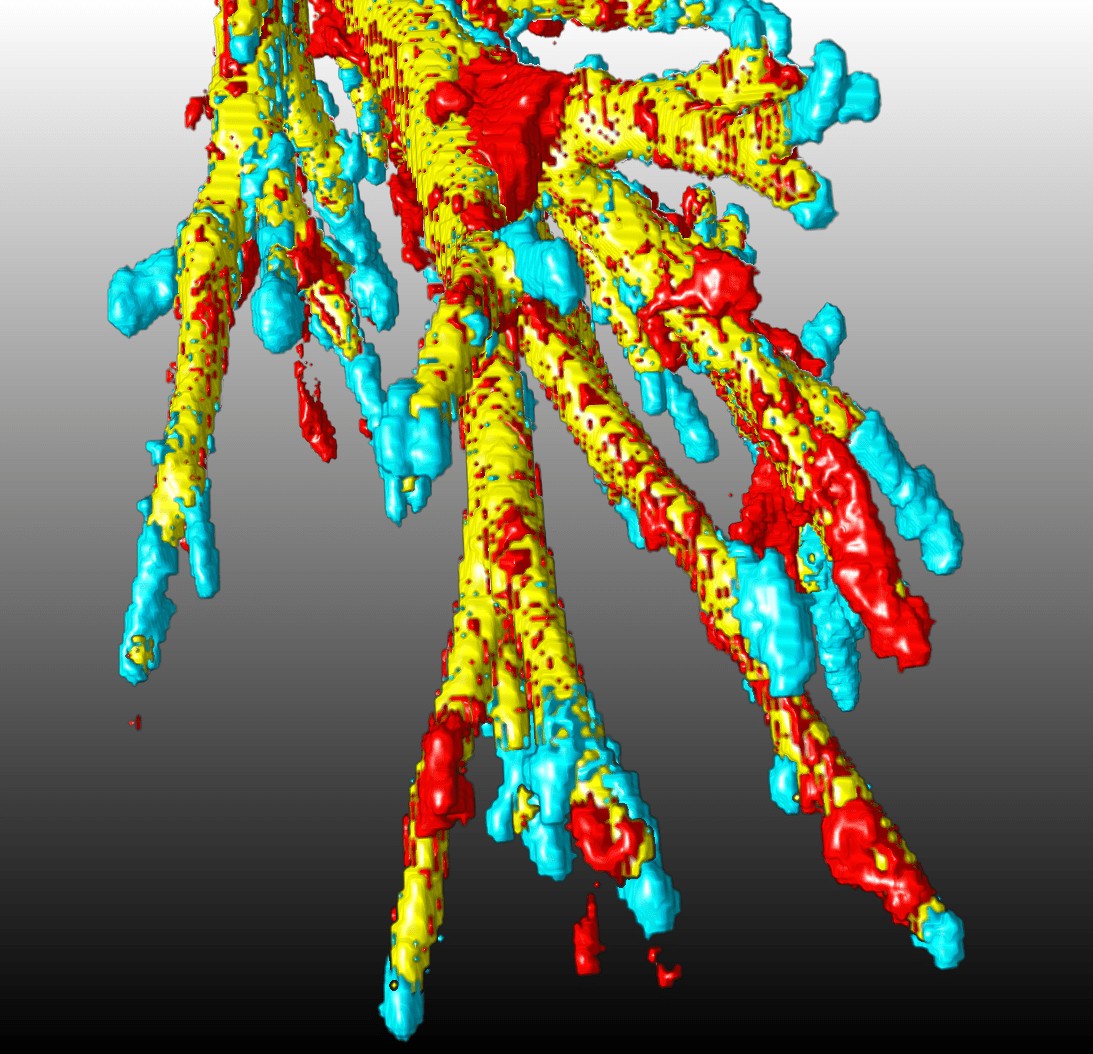}
\caption{wBCE-Rigid, detailed view}
\end{subfigure}
\centering
\begin{subfigure}{0.48\textwidth}
\includegraphics[width=\linewidth]{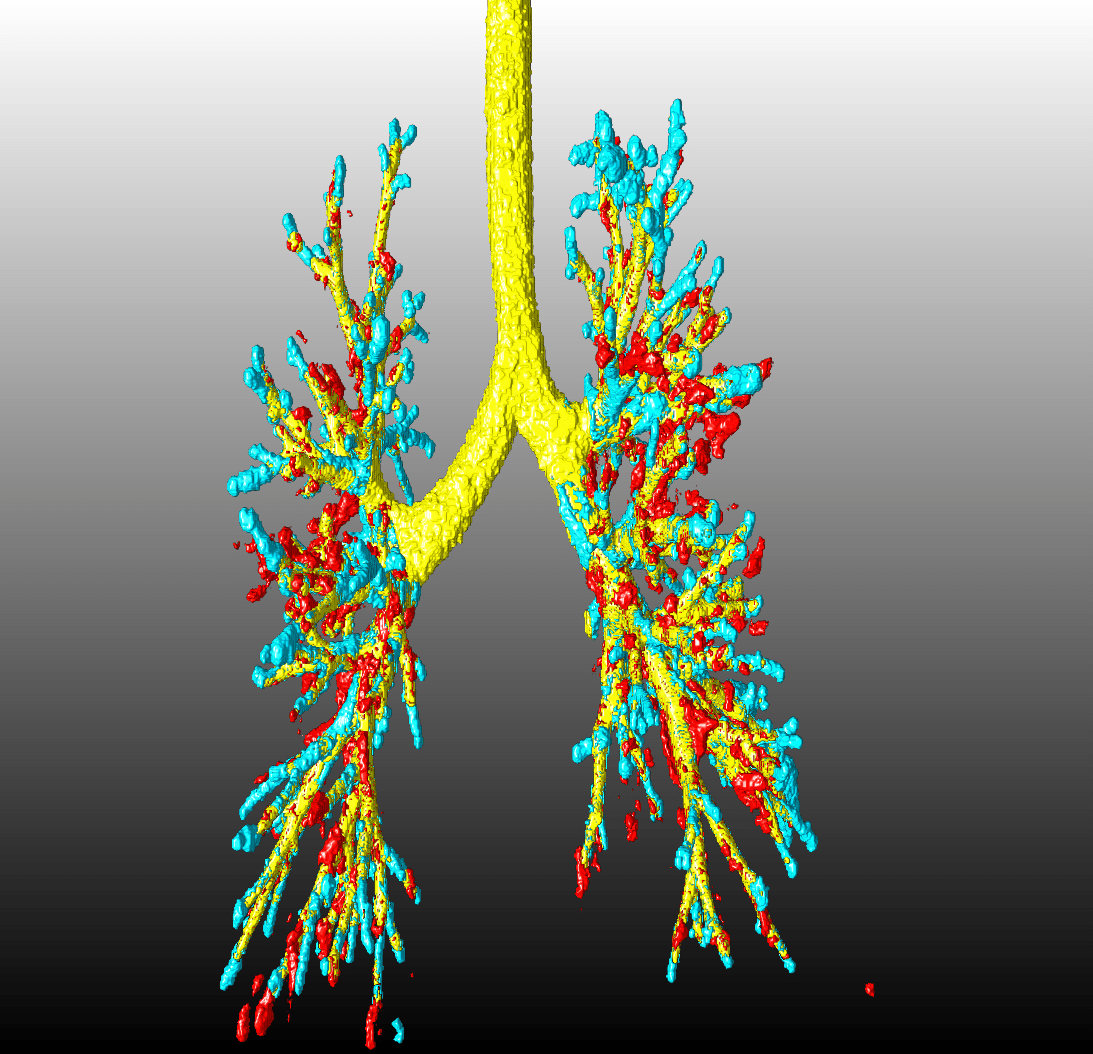}
\caption{dice-None}
\end{subfigure}
\centering
\begin{subfigure}{0.48\textwidth}
\includegraphics[width=\linewidth]{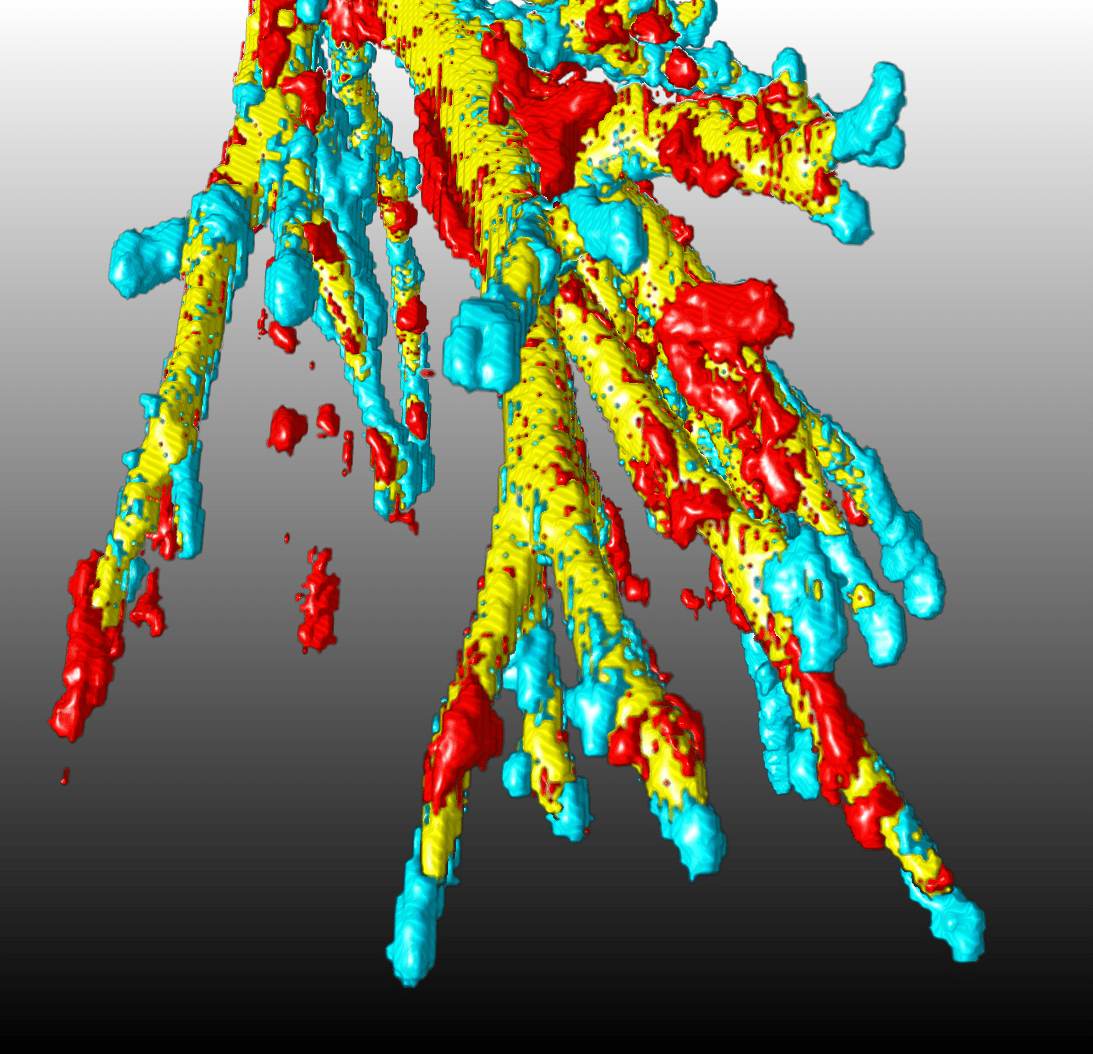}
\caption{dice-None, detailed view}
\end{subfigure}
\end{figure}
\begin{figure}\ContinuedFloat
\centering
\begin{subfigure}{0.48\textwidth}
\includegraphics[width=\linewidth]{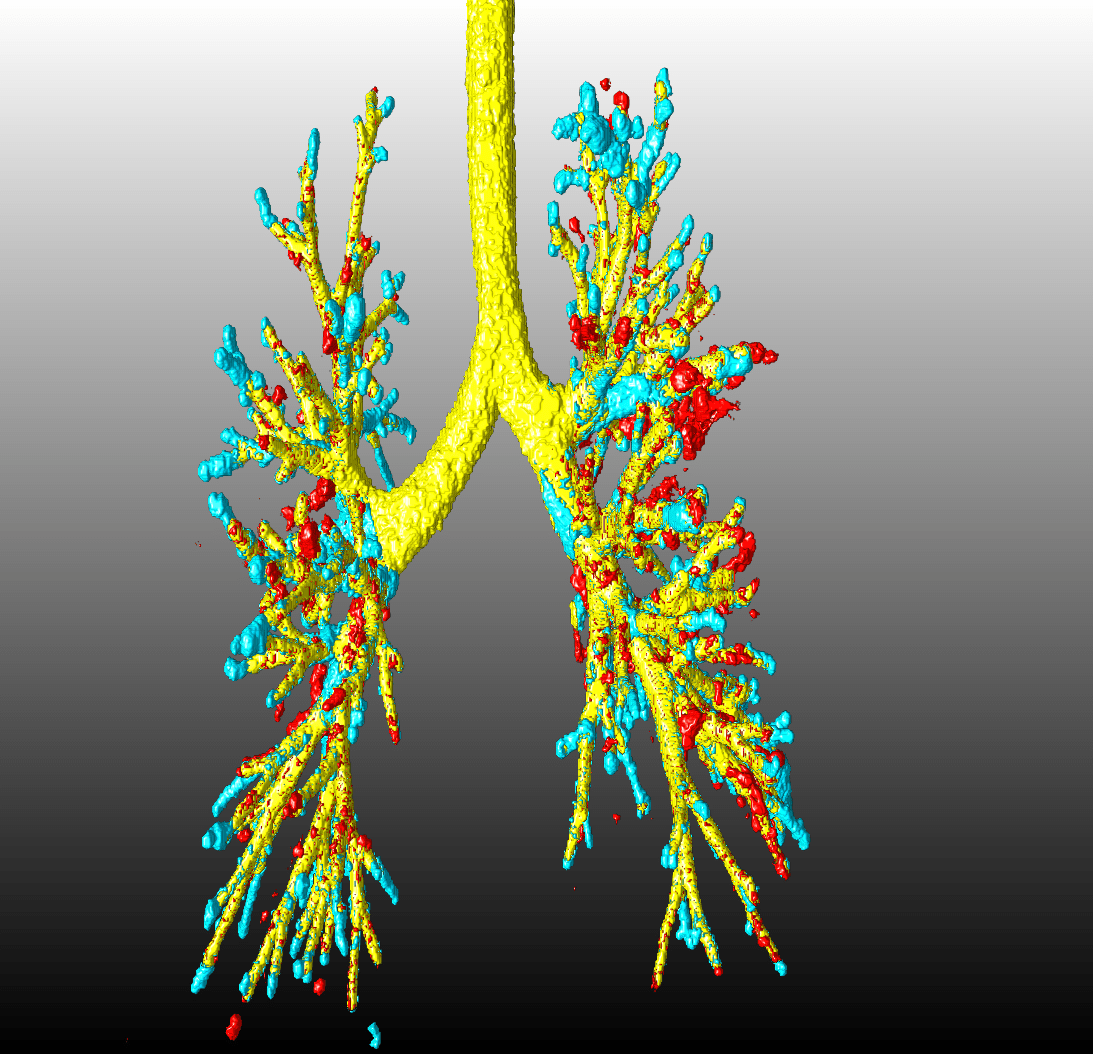}
\caption{dice-Rigid}
\end{subfigure}
\centering
\begin{subfigure}{0.48\textwidth}
\includegraphics[width=\linewidth]{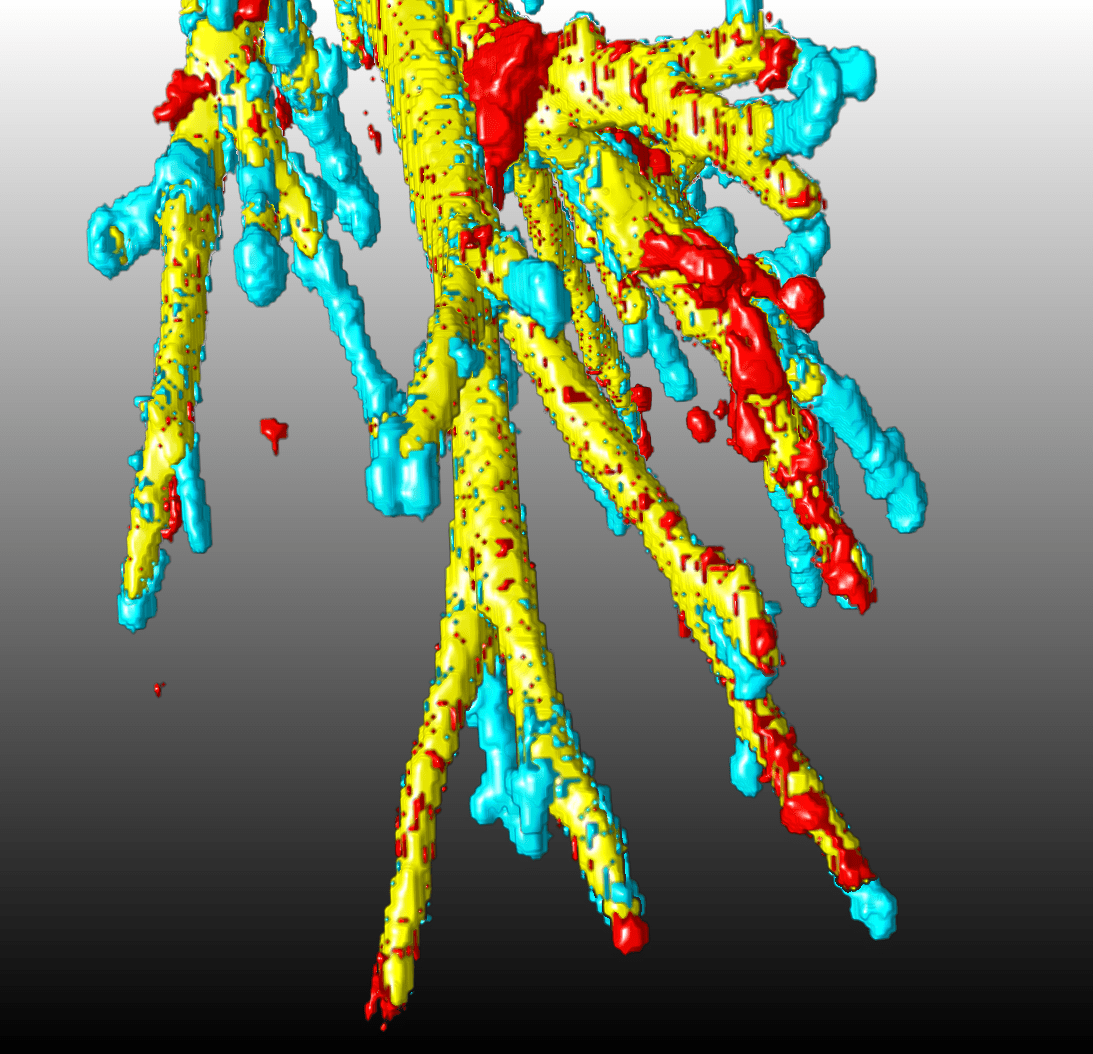}
\caption{dice-Rigid, detailed view}
\end{subfigure}
\centering
\begin{subfigure}{0.48\textwidth}
\includegraphics[width=\linewidth]{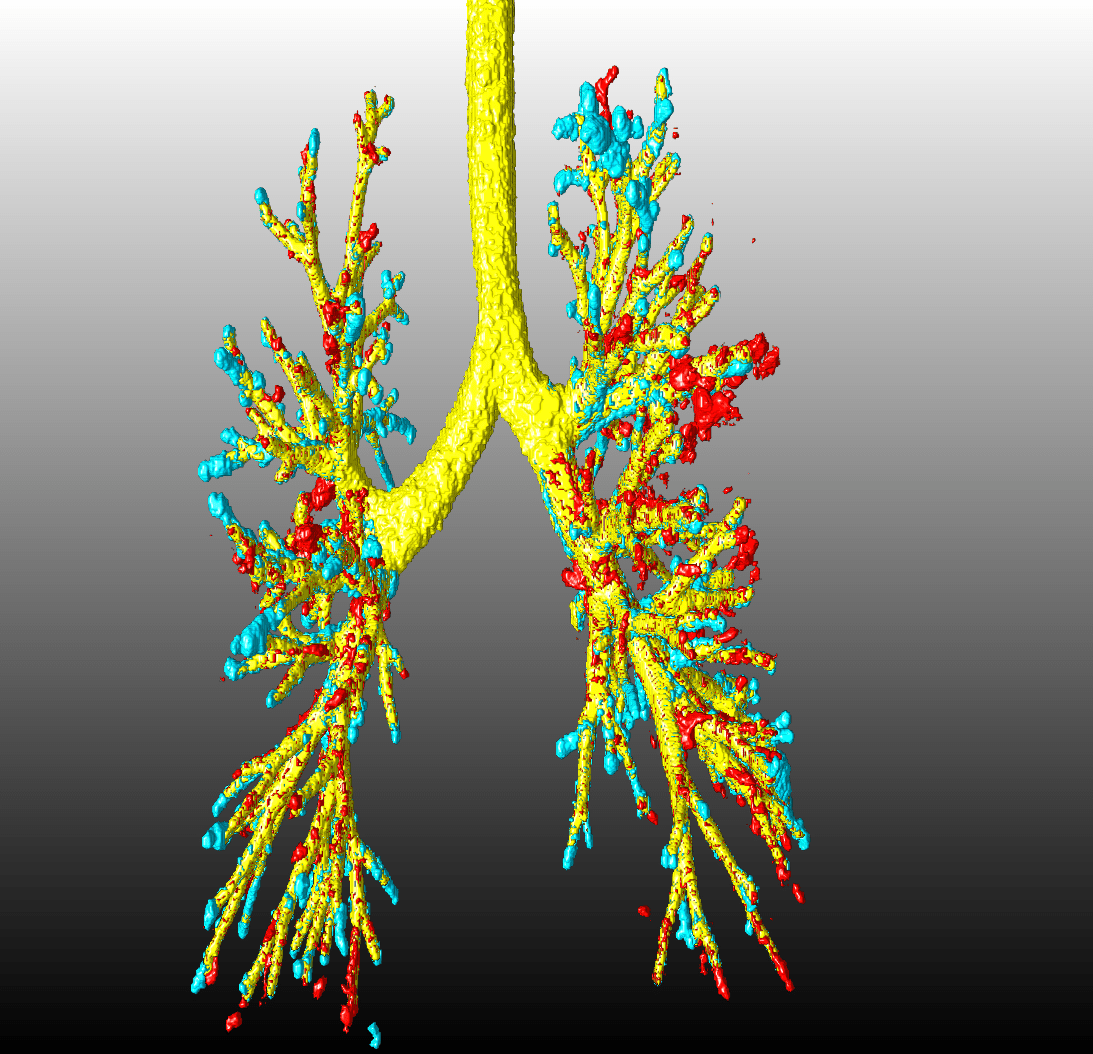}
\caption{dice-Elastic}
\end{subfigure}
\centering
\begin{subfigure}{0.48\textwidth}
\includegraphics[width=\linewidth]{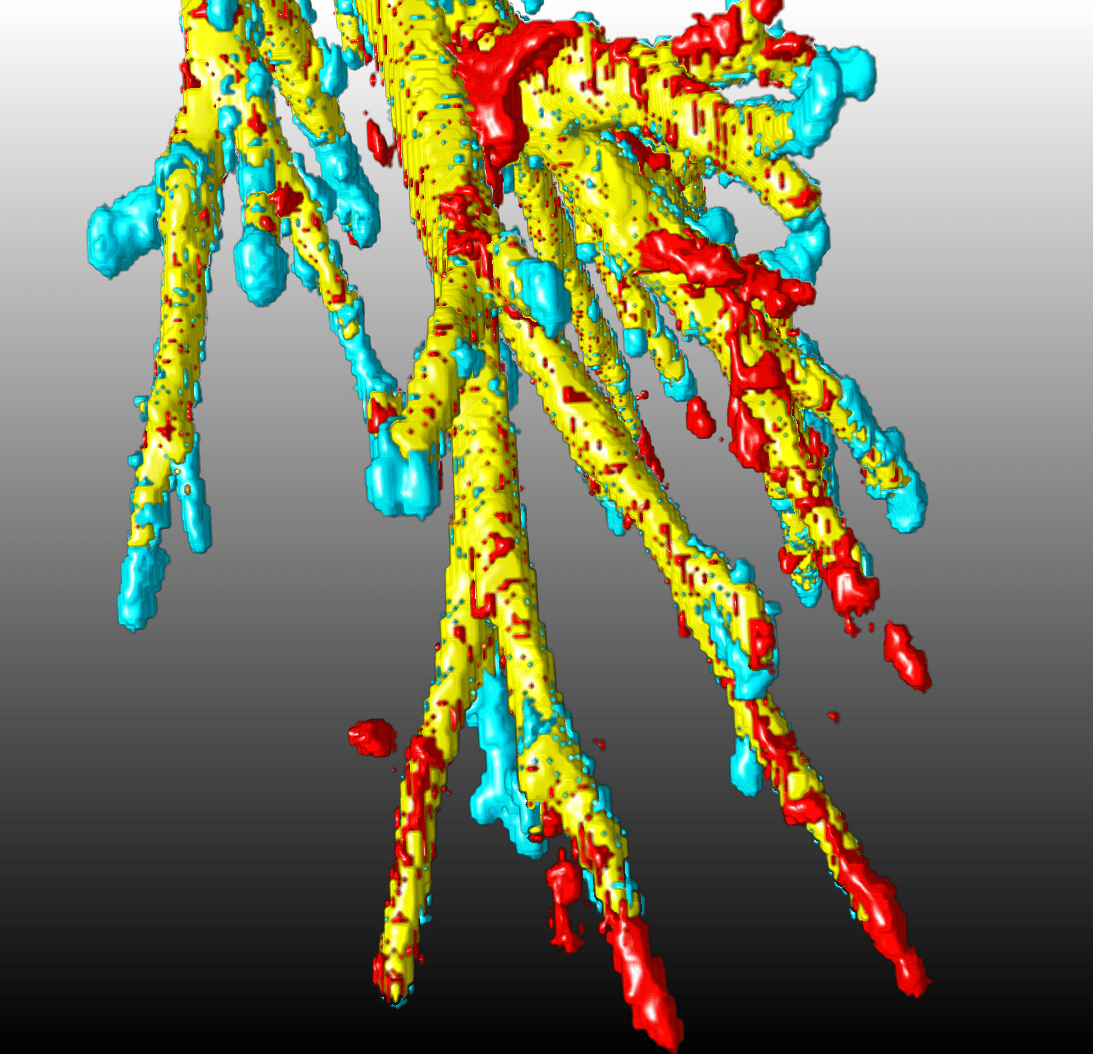}
\caption{dice-Elastic, detailed view}
\end{subfigure}
\caption{Left: visualization of false positives (red) and false negatives (blue) voxels, together with true positives (yellow), for the airway segmentations obtained for one of test CTs and all models tested (Table~\ref{tab1}). The results correspond to the optimal threshold for each model. Left: view of full airways tree. Right: detailed view around the peripheral branches in the lower-right section of the tree.} \label{fig5}
\end{figure}

The free ROC (FROC) curves for all models in Table~\ref{tab1} are displayed in Fig.~\ref{fig3}. This shows the sensitivity and number of false positives for the segmentations obtained when varying the threshold level in the probability maps. An optimal threshold value for each case is estimated as the point on the FROC curve closest to the upper-left corner.  The accuracy measured as the Dice coefficient averaged over the test set is also shown in Fig.~\ref{fig3}. The trachea and main bronchi are excluded from this calculation. For visualization, the segmented airway tree for one of test CTs obtained with the model "dice-Elastic" in Table~\ref{tab1} is displayed in Fig.~\ref{fig4}, together with the ground-truth. Furthermore, the false positives and false negatives voxels for the segmentations obtained with all models tested (Table~\ref{tab1}) on the same CT are displayed in Fig.~\ref{fig5}. These results correspond to the optimal threshold for each model.

The most accurate segmentations are obtained when using the dice loss function, and elastic deformations as data augmentation. The average Dice coefficient on the test set is 0.80, excluding the trachea and main bronchi. This accuracy is similar to the results reported in~\cite{ref8} on a larger dataset. 

In Fig.~\ref{fig5} it is seen that the largest errors are false negatives located in the tip of peripheral airways, which indicates that the method captures slightly shorter branches than the ground-truth. Also false positives are important in this region. Nevertheless, some of these errors might be due to missing smaller airways in the ground-truth, explained by the fact that some terminal branches were missed in the centrelines annotations. Other false positives are present in the form of small disconnected blobs. To reduce these errors, one could select the largest connected component or apply post-processing techniques such as Conditional Random Fields (CRF)~\cite{ref16}. Furthermore, when using wBCE loss, one could further reduce the number of misclassifications by locally increasing the weights at the peripheral airways.

In Figs.~\ref{fig3} and~\ref{fig5} it is shown how the use of data augmentation increases substantially the accuracy in the resulting segmentations, both when using wBCE and when using dice loss function. For either case, the average Dice increases approximately by 0.1 and 0.05, respectively.  The airways segmented with data augmentation show a much lower number of both false positives and false negatives. Also, they show more regular tubular structure, with uniform diameter along the branches. On the contrary, the results without data augmentation show branches with irregular blob shape, observed by the false positives around the branches. In particular, elastic deformation as data augmentation has been very efficient in our experiments, resulting in an increase of average Dice of approximately 0.05 when compared to the same set-up but using rigid transformations. Also, it is seen in Fig.~\ref{fig5} that it leads to further decrease in false negatives in peripheral branches. This agrees with the observation in~\cite{ref7} that "especially random elastic deformations of the training samples seem to be the key concept to train a segmentation network with very few annotated images".

It is shown in Figs.~\ref{fig3} and~\ref{fig5} that the use of dice loss function results in more accurate segmentation when compared to the wBCE loss. The tests with wBCE loss show over-segmented branches, as it is observed in Fig.~\ref{fig5} by the larger amount of false positives around the peripheral airways. This is due to the weighting for the airways class used in the wBCE formula. This compensates for the intraclass imbalance, but on the downside causes an over-segmentation of airways. In order to reduce this issue, one could adopt an approach where the ratio between the airways and background weights is reduced as the training of the network proceeds.
%
%
%\vspace{-0.2cm}
\section{Conclusions}
This paper shows a simple but robust method based on 3D Unets to perform segmentation of airways from chest CTs. Accurate segmentations have been obtained on a dataset containing 24 CTs, reaching a Dice coefficient of 0.8 between the ground-truth and our automated segmentations. Moreover, the importance of using data augmentation for our experiments has been demonstrated, in particular elastic deformations. In contrast to other CNNs-based methods, our approach is simpler and end-to-end optimised, and extracts a coherent and accurate airway tree based on voxelwise airway probabilities, with no need to input any prior knowledge of the connected tree structure.
\\
\\
\textbf{Acknowledgements.} This work has been funded by the EU Innovative Medicines Initiative (IMI). We would like to thank F. Dubost for his help with the experiments and with writing of this manuscript. We would also like to thank F. Calvet for sharing with us his implementation of elastic image deformation.

\end{document}